\documentclass[conference]{IEEEtran}
\IEEEoverridecommandlockouts
\usepackage{cite}
\usepackage{amsmath,amssymb,amsfonts}
\usepackage{algorithmic}
\usepackage{textcomp}
\usepackage{xcolor}

\usepackage{times}
\usepackage{soul}
\usepackage{url}
\usepackage[utf8]{inputenc}
\usepackage{graphicx}
\usepackage{booktabs}

\usepackage{ascmac}
\usepackage{listings}
\usepackage{subcaption}

\usepackage{multirow}

\definecolor{redcolor}{RGB}{255, 216, 216}
\definecolor{mycolor}{RGB}{216, 216, 255}

\usepackage[ruled, vlined, linesnumbered]{algorithm2e}
\usepackage{algorithmic}

\newcommand{\argmax}{\mathop{\rm arg~max}\limits}
\newcommand{\argmin}{\mathop{\rm arg~min}\limits}

\makeatletter
\newcommand{\figcaption}[1]{\def\@captype{figure}\caption{#1}}
\newcommand{\tblcaption}[1]{\def\@captype{table}\caption{#1}}
\makeatother

\newcommand{\graph}{\ensuremath{\mathcal{G}}}
\newcommand{\adjacency}{\ensuremath{\mathbf{S}}}
\newcommand{\attributes}{\ensuremath{\mathbf{X}}}
\newcommand{\labels}{\ensuremath{\mathbf{Y}}}
\newcommand{\emd}{\mathbf{z}}

\newcommand{\architecture}{\alpha}
\newcommand{\parameter}{\mathbf{W}}

\newtheorem{definition}{Definition}

\newcommand{\name}{ExGNAS\xspace} % problem name

\def\BibTeX{{\rm B\kern-.05em{\sc i\kern-.025em b}\kern-.08em
    T\kern-.1667em\lower.7ex\hbox{E}\kern-.125emX}}
\begin{document}

\title{Explainable Graph Neural Architecture Search via Monte-Carlo Tree Search (Full version)
}

\author{\IEEEauthorblockN{Yuya Sasaki}
\IEEEauthorblockA{
\textit{The University of Osaka}\\
Japan \\
sasaki@ist.osaka-u.ac.jp}
}

\maketitle

%\begin{abstract}
\begin{abstract}
The number of graph neural network (GNN) architectures has increased rapidly due to the growing adoption of graph analysis. Although we use GNNs in wide application scenarios, it is a laborious task to design/select optimal GNN architectures in diverse graphs. To reduce human efforts, graph neural architecture search (Graph NAS) has been used to search for a sub-optimal GNN architecture that combines existing components. However, existing Graph NAS methods lack explainability to understand the reasons why the model architecture is selected because they use complex search space and neural models to select architecture. Therefore, we propose an explainable Graph NAS method, called \name, which consists of (i) a simple search space that can adapt to various graphs and (ii) a search algorithm with Monte-Carlo tree that makes the decision process explainable. The combination of our search space and algorithm achieves finding accurate GNN models and the important functions within the search space. We comprehensively evaluate \name compared with four state-of-the-art Graph NAS methods in twelve graphs. Our experimental results show that \name{} achieves high average accuracy and efficiency; improving accuracy up to 26.1\% and reducing run time up to 88\%. Furthermore, we show the effectiveness of explainability by questionnaire-based user study and architecture analysis.
\end{abstract}
%\end{abstract}


\section{Introduction}
Graph  Neural Networks (GNNs) are powerful tools for practical data science tasks in various application scenarios from various domains including chemistry~\cite{fung2021benchmarking}, physics~\cite{sanchez2020learning}, and social science~\cite{kipf2017semi}.
Although we have developed numerous GNN architectures for various graphs, there are no one-size-fits-all GNN architectures yet.
It is laborious to design new GNN architectures and select optimal GNN architectures from numerous architectures according to the characteristics of graphs~\cite{kipf2017semi,maekawa2023simple,ma2021homophily,luan2021heterophily,zhu2020beyond,li2022finding,gasteiger2018predict,lim2021large,takahashi2025graph,maekawa2022beyond}.

To reduce human efforts and computational costs in designing/selecting GNN architectures, graph neural architecture search (Graph NAS) has been used to search for a sub-optimal GNN architecture in a given graph~\cite{ijcai2020p195,zhao2020simplifying,zhou2019auto}.
Due to the increasing number of GNN architectures and the demand for graph analysis, Graph NAS becomes important for researchers and practitioners.

\noindent
{\bf Issues in Graph NAS methods.}
Graph NAS methods have two technical design challenges; (1) search space and (2) search algorithm.
First, the search space defines patterns of GNN architectures; GNN architectures are generated by combinations of components within the search space. If the search space is not well-designed, there are no suitable GNN architectures within the search space.
Second, search algorithms determine how to preferentially search for GNN architectures that could achieve high accuracy.  
If search algorithms are not sophisticated, the search process becomes inefficient, and suitable GNN architectures are not found during the search process.

Existing Graph NAS methods address these two challenges to improve accuracy and efficiency.
However, their search algorithms use {\it neural models} as the backbone to run the search from {\it complex} search spaces. It causes an inefficient search process and a low capacity for analyzing the importance of components in GNN architectures.
In addition, their search spaces mainly focus on homophilic graphs, so they lack the adaptability to heterophilic graphs. 

\noindent
{\bf Motivation.}
We reconsider Graph NAS from the perspective of practical data science for researchers and practitioners.
For researchers, Graph NAS should provide (i) strong baselines on various graphs and (ii) the important components, which are helpful in designing new GNN architectures.
For practitioners, Graph NAS should efficiently and easily provide sub-optimal GNN architectures on various graphs without knowledge of GNNs, the characteristics of graphs, and heavy hyper-parameter tuning.
These indicate the necessity for efficient and explainable Graph NAS that provides sub-optimal architectures on various graph types.

Although explainable NAS methods have been studied widely~\cite{li2020random,real2020automl,ru2020interpretable}, there are no explainable Graph NAS methods, to the best of our knowledge.
For example, GraphGym~\cite{you2020design} aims to understand the important fundamental components in GNN architectures. It does not provide search algorithms and GNN architectures for heterophilic graphs.
AutoHeG~\cite{zheng2023auto} provides sub-optimal GNN architecture for heterophilic graphs, but it is inefficient and complex to understand the important components of GNN architectures. In addition, there are no studies that empirically compare the important components between homophilic and heterophilic graphs.
Therefore, we need to study efficient and explainable Graph NAS on homophilic and heterophilic graphs and analyze the important components across graph types.

\noindent
{\bf Contribution.}
We are the first to study explainable graph neural architecture search, to the best of our knowledge.
We propose an efficient and explainable Graph NAS method, called \name, which consists of a simple but effective search space and algorithm.
First, 
our search space is well-designed to achieve high accuracy in both homophilic and heterophilic graphs, despite being simple. Our search space includes fundamental components of GNN architectures such as multilayer perceptron (MLP), activation functions, and jumping knowledge networks~\cite{xu2018jumping}, while does not include any state-of-the-art GNN layers.
This leads to simple GNN architectures that can easily understand their components. 
Second, our search algorithm employs Monte-Carlo tree search without neural models, which makes the decision process explainable.
It selects the next GNN architecture from the average performance of explored architectures and finally outputs the Monte-Carlo tree with the importance of components. Our method does not require hyper-parameter tuning.
The elegant combination of the simple search space and algorithm achieves explainability, high efficiency, and adaptability on homophilic and heterophilic graphs.

We comprehensively evaluate our Graph NAS method compared with four Graph NAS methods in six homophilic and six heterophilic graphs.
Our experimental results show that \name achieves the highest average AUC. It improves accuracy up to 12.3 and reduces run time up to 88\% compared with the state-of-the-art Graph NAS methods.
Furthermore, we validate that \name helps to analyze the difference between GNN architectures in homophilic and heterophilic graphs.

Our contributions are summarized as follows:
(1) We first study an explainable graph neural architecture search,
(2) We propose an efficient and explainable graph neural architecture search method \name, and
(3) Extensive experimental studies demonstrate that \name{} outperforms the state-of-the-art baselines in heterophilic graphs and helps to analyze GNN architectures.

\noindent
{\bf Reproducibility.} We open our code at\\ \url{https://github.com/OnizukaLab/ExGNAS}.

\section{Preliminaries and Related work}
\label{sec:preliminary}

\subsection{Graphs}

An {\em undirected attributed graph with node labels} is a triple $\graph=(\adjacency,\attributes,\labels)$ where
$\adjacency \in \{0,1\}^{n\times n}$ is an adjacency matrix,
$\attributes \in R^{n\times d}$ is an attribute matrix assigning attributes to nodes, 
and a label matrix $\labels\in\{0,1\}^{n\times y}$ contains label of each node,
and $n$, $d$, and $y$ are the numbers of nodes, attributes, and labels, respectively.

\subsection{Neural Architecture Search}
Neural architecture search (NAS) aims to find the best model that achieves the highest accuracy in a given dataset.
A model is represented by a pair $(\architecture, \parameter)$ where $\architecture$ is a model architecture and $\parameter$ is a parameter of a neural network.
A search space $\mathcal{A}$ defines patterns of model architectures.
The best architecture $\architecture^*$ is defined as follows:

\begin{eqnarray}
\architecture^* = \argmax_{\architecture \in \mathcal{A}} {\rm E_{Val}} (\architecture, \parameter_{\architecture}^*).\\
\parameter_{\alpha}^* = \argmin_{\parameter} {\rm L_{Train}} (\architecture,\parameter).
\end{eqnarray}
\noindent
where $\rm{E_{Val}}$ and $\rm{L_{Train}}$ indicate the evaluation metrics on validation data and the loss function on train data, respectively. 

In this paper, we focus on node classification following existing studies~(e.g., 
\cite{guan2022large,li2020autograph,zhang2022deep,zhou2019auto}).
In a node classification, given a graph and partially labeled nodes, it predicts the labels of the rest of the nodes in the graph.

\subsection{Related Work}
\label{sec:related}

\noindent
{\bf Graph Neural Architecture Search}.
Graph NAS methods design their search space and search algorithms to find an optimal architecture in a given graph.
The recent survey summarizes the characteristics of Graph NAS methods~\cite{oloulade2021graph}.

\noindent
\underline{Search Space}.
The search space differs across studies. The search space itself can be considered a large technical contribution even if Graph NAS methods use existing search strategies~\cite{you2020design,zhang2022autogt}.
Most studies generate GNN layers by combining fundamental architectural components such as aggregation, attention, and activation functions, and connect the generated GNN layers by skip connection and JKNet~\cite{you2020design,li2020autograph,guan2022large}. 
GraphNAS~\cite{ijcai2020p195} uses functions such as activation and attention functions, and the number of heads, while the number of GNN layers is fixed. 
GraphGym~\cite{you2020design} includes pre-/post processing before/after GNN layers and skip connections.
AutoGraph~\cite{li2020autograph} focuses on skip connections and automatically selects the number of GNN layers.
Some methods (e.g.,  DFG-NAS~\cite{zhang2022deep,he2025decoupled}, NAS-Bench-Graph~\cite{qin2022bench}, and AutoHeG~\cite{zheng2023auto}) search for combinations of the state-of-the-art GNN layers with arbitrary patterns of connections.

\noindent
\underline{Search Algorithm}.
Search algorithms support efficiently finding the best GNN architecture within the search space.
The search algorithms have three representative approaches; reinforcement learning~\cite{zhou2019auto,ijcai2020p195,deng2023efficient}, evolutionary algorithm ~\cite{zhou2019auto,li2020autograph,zhang2022autogt}, and differentiable search~\cite{zheng2023auto,xu2023not,zhang2023dynamic,yao2024data,xie2024towards}.
A recent study uses large language models~\cite{wang2023graph} as search algorithms.
Monte-Carlo tree search belongs to reinforcement learning.
Existing reinforcement learning-based methods employ neural networks to maximize the expected performance of GNN models.
EGNAS~\cite{deng2023efficient} mixes Monte-Carlo tree search and deep reinforcement learning, which divides the search space by Monte-Carlo tree search and selects architectures by deep reinforcement learning. Thus, EGNAS does not use Monte-Carlo tree search to select architectures.
There are no Graph NAS methods that only use Monte-Carlo tree search.

\noindent
{\bf Explainable NAS}.
Explainable NAS methods are actively studied to explain their decision process. 
Their common design rationale is to avoid using neural models~\cite{carmichael2021learning,blazek2021explainable,liu2021fox,ru2020interpretable}.
FOX-NAS uses a simulated annealing~\cite{liu2021fox}, and NAS-BOWL uses Bayesian optimization with Weisfeiler-Lehman kernels~\cite{ru2020interpretable}.
XNAS~\cite{carmichael2021learning} employs evolutionary algorithms.
He~et al.~\cite{he2025decoupled} uses mean distinguishability to evaluate how to distinguish node representation on GNNs.
There are no NAS methods with Monte-Carlo tree search for explainability yet.

\section{\name: Explainable Graph NAS}

In this section, we present our graph neural architecture search method, called \name.
We first define our problem. 

\begin{definition}[Explainable Graph NAS Problem]
Given graph $\graph$, train/val/test datasets, and search space $\mathcal{A}$, the explainable graph neural architecture search problem aims to output (1) the model that achieves the highest classification performance with explainable decision process and (2) the importance of components in $\mathcal{A}$. 
\end{definition}

The definition of the explanation follows the reasoning and behavior defined by Nauta~et al.~\cite{nauta2023anecdotal}.
Our explanation aims to know how a method finds a particular model and how a method operates without analyzing the internal workings by observing input and output.

\subsection{Search Space}
\begin{figure}[ttt]
     \centering
     \includegraphics[width=0.6\linewidth]{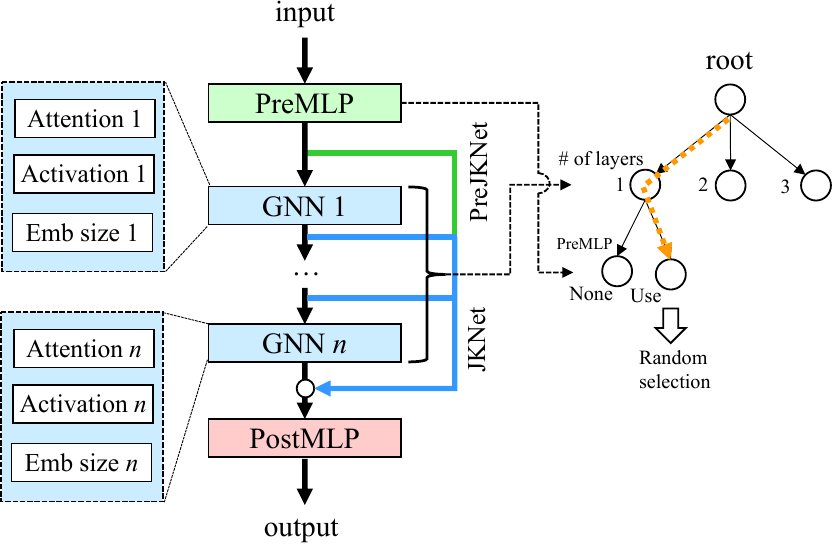}
       \figcaption{Search space and Monte-Carlo tree search; each architectural component and parameter are assigned to tree depth and node in Monte-Carlo tree, respectively.}
       \label{fig:designspace}
\end{figure}

The search space defines patterns of GNN architectures.
If the search space is well-designed, Graph NAS methods can cover sub-optimal models for diverse graphs.
Our search space includes two aspects of GNN architectures: Micro-architecture and macro-architecture. 
The micro-architecture specifies the inside of independent GNN layers, and the macro-architecture specifies the connections between GNN layers.
The left-side image in Figure~\ref{fig:designspace} illustrates our search space on GNN architectures. Intuitively, GNN~$1$--GNN~$n$ indicate the micro-architectures and the whole structure indicates the macro-architecture.

\noindent
{\bf Design policy}.
We design the search space that includes fundamental components without any state-of-the-art complex techniques.
This search space generates non-complex GNN architectures that help to understand what components are effective in the given graph. 
To adapt to heterophilic graphs, our search space includes components for emphasizing their own node attributes and jumping knowledge which are known to be effective in heterophilic graphs~\cite{abu2019mixhop,chien2021adaptive,zhu2020beyond}.

We represent our search space by a set of architecture parameters instead of supernets~\cite{liu2018darts}.
Supernets are often used for NAS, in particular computer vision tasks. A set of architecture parameters is helpful in analyzing the important components because it is easy to compare the generated architectures, while supernets are hard to understand the important components due to complex architectures.  

\noindent
{\bf Micro-architecture}.
Multi-layer message passing model is a standard GNN architecture.
Message-passing models learn feature representations of nodes over layers.
Formally, the $l$-th GNN layer can be defined as:

\begin{equation}
\emd_u^{(l)} = \sigma \left( \sum_{v\in \mathcal{N}_u} e_{(u,v)} \parameter^{(l)} \emd_v^{(l-1)} \right).
\end{equation}
\noindent
where $\emd_u^{(l)}$ is the $l$-th layer embedding of node $u$, $\parameter^{(l)}\in R^{|\emd^{(l)}|\times |\emd^{(l-1)}|}$ is trainable weights, and $\mathcal{N}_u$ is the neighborhood of node $u$. $\emd_u^{(0)}$ is the attribute of node $u$.
$e_{(u,v)}$ and $\sigma$ indicate the attention between nodes $u$ and $v$ and activation function, respectively.

Our micro-architecture has three design dimensions; attention, activation, and embedding size.
These are common and standard components in GNN layers.
It does not include model-specific dimensions (e.g., the number of heads for GAT) to reduce the number of GNN architecture patterns.

\noindent
{\bf Macro-architecture}.
The macro-architecture determines how GNN layers are organized into a whole neural network.
The common way is to stack multiple GNN layers. 
We use jumping knowledge network ({\it JKNet}) that connects each output of GNN layer to the output of the final GNN layer.
We also use multilayer perceptron (MLP) layers before/after GNN layers; we call them {\it preMLP} and {\it postMLP}, respectively.
In addition, we add skip connections from preMLP to the final outputs; we call it {\it PreJKNet}.

This macro-architecture benefits achieving high accuracy in heterophilic graphs.
It is well-known that in heterophilic graphs, node attributes are often more important than feature aggregation from neighborhoods because neighborhoods often have different labels~\cite{ma2021homophily,luan2021heterophily,you2020design,zhu2020beyond}.
The combination of PreMLP and PreJKNet could be effective for heterophilic graphs, though no Graph NAS studies investigate their effectiveness.

\noindent
{\bf Architecture parameters}. 
We show the architecture parameters of the search space in
Table~\ref{tab:searchspace}.
We separately set architecture parameters for each GNN layer, for example, when the number of GNN layers is two, we can set the embedding sizes of the first and second GNN layers are 32 and 128, respectively.
The search space includes over 20 million GNN architecture patterns in total.
We note that there are dependencies of some architecture parameters, for example, if JKNet is ``max'', all embedding sizes of PreMLP and GNN layers should be the same.
We can control the embedding sizes to be fitted to graph size and GPU memory size adaptively.

\begin{table}[ttt]
    \centering
    \caption{Summary of architecture parameters. $d_u$ and $||$ indicate the degree of node $u$ and concatenation operation, respectively.}
    \label{tab:searchspace}
    \scalebox{0.85}{
    \begin{tabular}{|l|l|}\hline
         Component  &\multicolumn{1}{c|}{Parameter}  \\\hline\hline
         The number of GNN layers & 1, 2, 3 \\\hline
          \multirow{3}{*}{Attention} & Constant: $e_{(u,v)}=1$ \\
         & GCN: $e_{(u,v)}=1/\sqrt{d_u d_v}$ \\
         & GAT: $e_{(u,v)}=leakyReLU(\parameter_l \emd_u || \parameter_r \emd_v)$ \\\hline
         Activation & None, Relu, Sigmoid, Tanh \\\hline
         The embedding size of&\multirow{2}{*}{ 16, 32, 64, 128, 256, $y$} \\
         of GNN layer & \\\hline
         JKNet& None, concat, max \\\hline
         PreJKNet& None, use \\\hline
         PreMLP & None, use \\\hline
         The embedding size &\multirow{2}{*}{16, 32, 64, 128, 256} \\
         of preMLP &\\\hline
         The number of layers & \multirow{2}{*}{None, 1, 2} \\
         in postMLP & \\\hline
         The hidden size& \multirow{2}{*}{64, 128, 256} \\
         of postMLP &  \\\hline
    \end{tabular}
    }
\end{table}

\noindent
{\bf Difference between search spaces in existing and our studies}. 
Each existing study uses its own search space, such as aggregation in GNN and pre- and post-process MLPs.
Our search space has two characteristics.
First, we eliminate complicated GNN methods (e.g., H2GCN) from our search space to generate models with simple functions: improving the interpretability of the effectiveness of found models.
Second, our search space includes functions for non-homophily graphs, such as jumping knowledge (and pre-jumping knowledge) and Pre-MLP. 
We aim to handle both homophilic and heterophilic graphs in a single search space.

\subsection{Search algorithm}
The characteristics of our search algorithm are using a Monte-Carlo search tree and not using neural models, which makes the search process explainable and efficient.
This approach is simple but effective in efficiently selecting GNN architectures.

The right-side image in Figure~\ref{fig:designspace} illustrates how to select architecture parameters via Monte-Carlo tree search.
It traverses the Monte-Carlo tree to fix the architecture parameters.  
We assign components in the search space to tree depths and their architecture parameters to nodes on Monte-Carlo tree (we call node on Monte-Carlo tree {\it MCT node}). For example, in this figure, the number of layers is assigned to the first layer. 

\noindent
{\bf Design policy}.
Search algorithms often employ neural models, such as differentiable algorithms (e.g.~\cite{liu2018darts}) and deep reinforcement learning~(e.g.~\cite{ijcai2020p195}), to select the next model architectures.
Neural model-based algorithms could be effective in preferentially selecting highly accurate GNN architectures.
However, it needs a learning process for search algorithms which leads to inefficiency and inexplainability.
Therefore, we employ Monte-Carlo tree search without neural models, which makes the decision process explainable and can output the importance of components. In addition, Monte-Carlo tree search is more scalable (i.e., less memory usage) than other reinforcement learning such as Q-learning~\cite{q-learning} because Monte-Carlo tree search only manages the constructed tree instead of whole tables like Q-learning.

\noindent
{\bf Architecture selection}.
The process to select a GNN architecture is as follows:
(1) select a leaf MCT node, (2) fix architecture parameters according to MCT nodes on the path from the root to the selected leaf node, and (3) randomly fix other architecture parameters that are not fixed yet.

To select leaf MCT node $i$, we define a score of MCT node, which is an extension of UCB (Upper Confidence Bound)~\cite{kocsis2006bandit}, to find architectures that are expected to have high accuracy.
We select leaf MCT node $i$ that has the maximum $ucb$ by the following:
\begin{equation}
    ucb(i) =\frac{\sum_{(\architecture, \parameter) \in \mathcal{M}_i} {\rm E_{{Val}}}(\architecture, \parameter)}{m_i} + c\sqrt{\frac{\ln M}{m_i}} 
\end{equation}
\noindent
where $\mathcal{M}_i$ is a set of evaluated models with architecture parameters of node $i$. 
$m_i$ indicates the selected times of $i$ and its descendants, and $M$ indicates the number of explored models.
$c$ is a constant to control the balance of exploration and exploitation; larger values of $c$ correspond to larger amounts of exploration.
Since the value of $ucb(i)$ decreases as $m_i$ increases, Monte-Carlo tree search can search for globally optimal rather than locally optimal GNN architectures.

\noindent
{\bf Tree update}.
After selecting model architectures, it trains and tests models to validate the performance of selected GNN architectures.
Then, it updates $ucb$ of MCT nodes on the path from the root to the selected MCT node. If the MCT node $i$ is selected at $\theta$ times, it generates its child nodes; these child nodes are preferentially selected next time because $m=0$ (i.e., $ucb = \inf$).
We keep the average accuracy performance, the number of times selected, and the average training time on MCT nodes to understand how components affect accuracy and efficiency. 

\noindent
{\bf Hyper-parameters}.
\name has two hyper-parameters $c$ and $\theta$. We use $\sqrt{2}$ and 10 as default $c$ and $\theta$, respectively, following existing study~\cite{kocsis2006bandit}. We show that these values have a small impact on the accuracy performance in experimental studies.
Thus, \name does not require hyper-parameter tuning.

\noindent
{\bf Component order}.
We determine the order of components to effectively find the best architecture.
Since the components assigned to small tree depths are fixed earlier, we can investigate the importance of such components. We design component orders according to the impact on the performance following the trends and existing studies.

We first fix the number of GNN layers, preMLP, preJKNet, and JKNet because they can be considered as the components that highly affect the performance ~\cite{ma2021homophily,luan2021heterophily,zhu2020beyond,li2022finding,gasteiger2018predict,lim2021large}.
Then, we fix activation and attention functions in the first GNN layer, the embedding size of PreMLP, the hidden unit size of PostMLP, and the embedding size of the first GNN layer in this order.
Finally, we fix the other components; activation function, attention function, and the embedding size of later GNN layers, which are hard to expect their importance.

\noindent
{\bf Explainability}.
The Monte-Carlo tree search selects GNN architectures according to $ucb$ scores of MCT nodes computed from the accuracy performance and the number of selected times.
Therefore, we can know the reason why the next GNN architectures are selected.
We keep the average node classification performance and the average training time on MCT nodes to understand how components affect accuracy and efficiency. 
As \name{} outputs the Monte-Carlo tree as well as the best GNN architecture, we can know the importance of components.

\noindent
{\bf Pseudo-code}.
Algorithm~\ref{alg:bis} shows the pseudo-code of \name.
It repeatedly selects architecture $\architecture$ and trains the model (lines 3--6).
After training models, it updates the Monte-Carlo tree according to the accuracy of validation data (lines 7--9).
It updates the best model $\architecture^*$ and $\parameter^*$ if finding better models (lines 10--12).
It repeats these procedures $L$ times (line 2).

\noindent
{\bf Time complexity}.
The architecture selection process traverses Monte-Carlo tree from the root to leaf nodes, and then randomly selects other parameters.
It takes $O(|\mathbb{F}||\mathbb{A}_m|)$ where $|\mathbb{F}|$ and $|\mathbb{A}_m|$ are the number of components and the maximum number of parameters among components, respectively.
$|\mathbb{F}|$ and $|\mathbb{A}_m|$ correspond to the maximum tree depth and the maximum number of child nodes, respectively.

Our search algorithm repeated the architecture selection and tree update $L$ times, which is a given number of explored GNN architectures.
Consequently, the whole time complexity is $O(L |\mathbb{F}||\mathbb{A}_m| t_{\alpha})$, where $t_{\alpha}$ indicates the training time of model $\alpha$.
Typically, $|\mathbb{F}|$ and $|\mathbb{A}_m|$ are quite small, so the run time of \name is dominated by $L$ and model training.

% \begin{algorithm}[!t] 
% \caption{\name}	\label{alg:bis}
% \begin{algorithmic}[1]\vspace{-5mm}
% %\begin{multicols}{2}\vspace{-10mm}
% \Require {Graph \graph, \# of architectures $L$}
% \State Initialize $MCT$, $best$;
% \For{$1, \ldots, L$}
% \State $i \leftarrow$ leaf MCT node with the max $ucb$;
% \State $\architecture \leftarrow$ architecture of $i$;
% \State Initialize $\parameter$ of $\architecture$;
% \State Train model $(\architecture, \parameter)$ on \graph;
% \State Update $MCT$; $m_i \leftarrow m_i+1$;
% \If{$m_i \geq \theta$}
% \State Generate child MCT nodes of $i$;
% \EndIf         
% \If{$best<\rm{E_{{VAL}}(\architecture, \parameter)}$}
% \State $\architecture^*,\parameter^* \leftarrow \architecture, \parameter$;
% \State $best \leftarrow \rm{E_{{VAL}}(\architecture, \parameter)}$;
% \EndIf 
% \EndFor
% %\State Initialize $MCT$, $best$
% %\State {\bf return} $\architecture^*$, $\parameter^*$, $MCT$
% %\end{multicols}
% \vspace{-4mm}
% \end{algorithmic}
% \end{algorithm}

\begin{algorithm}[!t] 
%\begin{multicols}{2}
	\caption{\name}	\label{alg:bis}
         \SetKwInOut{Input}{input}
        \SetKwInOut{Output}{output}
        \Input{\graph, $L$}
        \Output{$\architecture^*$, $\parameter^*$, $MCT$}
	%{\small
	    Initialize $MCT$, $best$;\\
            \For{$1, \ldots, L$}{
                $i \leftarrow$ leaf MCT node with the maximum $ucb$;\\
                $\architecture \leftarrow$ functions of $n_i$ with random parameters;\\
                Initialize $\parameter$ of $\architecture$;\\
                Train model $(\architecture, \parameter)$ on \graph;\\
                Update $MCT$;\\
                \If{$m_i \geq \theta$}{
                    Generate child MCT nodes of $i$;\\
                }
                \If{$best<\rm{E_{{VAL}}(\architecture, \parameter)}$}{
                    $\architecture^*,\parameter^* \leftarrow \architecture, \parameter$;\\
                    $best \leftarrow \rm{E_{{VAL}}(\architecture, \parameter)}$;\\
                }
%        \EndFor
             }
          {\bf return} $\architecture^*$, $\parameter^*$, $MCT$;\\
          {\bf end procedure}
%\end{multicols}
  %  }
\end{algorithm}

\section{Experimental study}
\label{sec:experiment}
We present the results of an experimental evaluation of our method on node classification tasks.
We designed the experiments for (1) evaluating the performance of \name in terms of classification performance, efficiency, and model sizes, (2) analyzing \name, and (3) analyzing the explainability and found architectures.
We implemented our algorithms in Python3 and used a server with NVIDIA V100 Tensor Core GPU and 16 GB GPU memory which is provided as instance p3.x2large on Amazon Web Service.

\subsection{Experimental Setting}

\noindent
{\bf Dataset}.
We use twelve graphs that are commonly used in GNN tasks (see Table~\ref{tab:overview_acc}).
These graphs include several application domains with different degrees of edge homophily~\cite{zhu2020beyond}.
Table~\ref{tab:dataset} shows the statistics of graphs.
%are in our code.

\begin{table}[ttt]
    \centering
    \caption{Dataset summary}
    \label{tab:dataset}
%     \vspace{-3mm}
    \scalebox{0.8}{
    \begin{tabular}{c|r|r|r|r|r}\midrule
         Dataset  &\multicolumn{1}{c|}{\# nodes} & \multicolumn{1}{|c|}{\# edges} & \multicolumn{1}{|c}{\# features}& \multicolumn{1}{|c|}{\# labels}&\multicolumn{1}{|c}{Edge homophily}\\ \midrule
         {\bf Cora} & 2{,}708 & 5{,}429 & 1{,}433 & 7 & 0.81\\
         {\bf CiteSeer} & 3{,}327 & 4{,}732 & 3{,}703 & 6 & 0.73\\
         {\bf Amz-P} & 7{,}650 & 238{,}162 & 745 & 8 & 0.82\\
         {\bf Amz-C} & 13{,}752 & 491{,}722 & 767 & 10 & 0.77\\
         {\bf Co-CS} & 18{,}333 & 163{,}788 & 6{,}805 & 15 & 0.79\\         
         {\bf PubMed} & 19{,}717 & 44{,}338 & 500 & 3 & 0.81\\\midrule
         {\bf Cornell} & 195 & 304 & 1{,}703 & 5 & 0.13\\
         {\bf Wisconsin} & 265 & 530 & 1{,}703 & 5 & 0.20\\
         {\bf Chameleon} & 2{,}277 & 36{,}101 & 2{,}325 & 5 & 0.23\\
         {\bf Squirrel} & 5{,}201 & 217{,}073 & 2{,}089 & 5 & 0.23\\
          {\bf Actor} & 7{,}600 & 30{,}019 & 932 & 5 &0.23\\
         {\bf Penn94} & 38{,}815 & 2{,}498{,}498 & 4{,}772 & 2 & 0.53\\
         \midrule
    \end{tabular}
    }
\end{table}

\begin{table*}[ttt]
    \centering
    \caption{Overview of Accuracy in our methods and baselines with variances. The best and worst results per dataset are highlighted by blue and red, respectively. DNF stands for `Did Not Finish within 48 hours,' and OOM refers to `Out of Memory.' }
    \label{tab:overview_acc}
%    \vspace{-2mm}
\begin{minipage}[t]{1.0\linewidth}
  \centering
%  \vspace{-2mm}
    \scalebox{0.95}{
    \begin{tabular}{crrrrrr|rr}\midrule
            &\multicolumn{8}{c}{{\bf Heterophilic}}\\
           &         {\bf Cornell} &              {\bf Wisconsin} &          {\bf Chameleon} &     {\bf Squirrel} &    {\bf Actor}&  \multicolumn{1}{r}{{\bf Penn94}} & \multicolumn{1}{|c}{Avg.}&\multicolumn{1}{c}{Rank}\\\midrule
           GraphNAS & \colorbox{redcolor}{69.2$_{\pm 0.3 }$}& \colorbox{redcolor}{74.7$_{\pm 0.2 }$}& 57.5$_{\pm 0.1 }$& \colorbox{redcolor}{36.7$_{\pm 0.7 }$}& \colorbox{redcolor}{33.2$_{\pm 0.0}$}& DNF &\colorbox{redcolor}{54.3} & \colorbox{redcolor}{4.3}\\
           GraphGym & 74.9$_{\pm 5.9 }$& 87.2$_{\pm 6.3 }$& \colorbox{redcolor}{54.0$_{\pm 2.0}$} & 37.2$_{\pm 1.3}$ & \colorbox{mycolor}{{\bf 38.3$_{\pm 1.0}$}}& \colorbox{mycolor}{{\bf 85.6$_{\pm 0.4}$}} &62.9& 2.2\\
           DFG-NAS & \colorbox{mycolor}{{\bf 83.1$_{\pm 0.4}$}}& \colorbox{mycolor}{{\bf 91.7$_{\pm 0.1}$}}& 67.0$_{\pm 0.1}$ & 47.2$_{\pm 0.0}$ & 38.1$_{\pm 0.0 }$& \colorbox{redcolor}{76.1$_{\pm 0.0 }$}&67.2& \colorbox{mycolor}{{\bf 1.8}}\\
           Auto-HeG & 74.9$_{\pm 14.2 }$& 86.8$_{\pm 39.9 }$& OOM & OOM & 38.0$_{\pm 1.0}$ & OOM &66.6& 3.7\\\midrule
            \name &  72.3$_{\pm 0.3 }$& 84.2$_{\pm 0.1 }$& \colorbox{mycolor}{{\bf 69.9$_{\pm 0.0}$}}& \colorbox{mycolor}{{\bf 59.5$_{\pm 0.0}$}}& 37.9$_{\pm 0.0 }$& 82.5$_{\pm 0.0}$ & \colorbox{mycolor}{{\bf 67.7}}& 2.7\\\midrule
    \end{tabular}
    }
\end{minipage}
\begin{minipage}[t]{1.0\linewidth}
  \centering
%  \vspace{-2mm}
    \scalebox{0.95}{
    \begin{tabular}{crrrrrr|rr}\midrule
            &\multicolumn{8}{c}{{\bf Homophilic}}\\
           & {\bf Cora} &         {\bf CiteSeer} &        {\bf Amz-P}  &   {\bf Amz-C} &         {\bf Co-CS} &        \multicolumn{1}{r}{{\bf PubMed}} & \multicolumn{1}{|c}{Avg.}&\multicolumn{1}{c}{Rank}\\\midrule
           GraphNAS  & 87.1$_{\pm 0.0}$ & 76.0$_{\pm 0.0}$ &95.8$_{\pm 0.0}$ & \colorbox{mycolor}{ \bf{91.9$_{\pm 0.0}$}} & \colorbox{redcolor}{94.2$_{\pm 0.0}$} & 87.8$_{\pm 0.0}$&88.8& 2.7\\
           GraphGym & 82.4$_{\pm 1.6}$ & 74.2$_{\pm 1.5}$ & \colorbox{mycolor}{ \bf{96.2$_{\pm 0.3}$}} & \colorbox{mycolor}{ \bf{91.9$_{\pm 0.3}$}} & 95.3$_{\pm 0.4}$ & \colorbox{mycolor}{ \bf{89.7$_{\pm 0.7}$}}&88.3& \colorbox{mycolor}{{\bf 2.3}}\\
           DFG-NAS & \colorbox{mycolor}{ \bf{88.5$_{\pm 0.0}$}} & \colorbox{mycolor}{ \bf{77.1$_{\pm 0.0}$}} & \colorbox{redcolor}{95.2$_{\pm 0.0}$} & \colorbox{redcolor}{88.0$_{\pm 0.0}$} & \colorbox{mycolor}{ \bf{95.8$_{\pm 0.0}$}} & \colorbox{redcolor}{87.6$_{\pm 0.0}$}&88.7& 2.6\\
           Auto-HeG& \colorbox{redcolor}{75.5$_{\pm 19.9}$} & \colorbox{redcolor}{70.5$_{\pm 3.5}$} & OOM & OOM & OOM & 88.3$_{\pm 0.1}$&\colorbox{redcolor}{78.1}& \colorbox{redcolor}{4.7}\\\midrule
            \name &  \colorbox{mycolor}{\bf{88.5$_{\pm 0.0}$}} & {74.7$_{\pm 0.0}$} & {95.6$_{\pm 0.0}$} & 91.2$_{\pm 0.0}$ & {95.6$_{\pm 0.0}$} & {89.3$_{\pm 0.0}$} &\colorbox{mycolor}{{\bf 89.2}}& \colorbox{mycolor}{{\bf 2.3}}\\\midrule
    \end{tabular}
    }
\end{minipage}
\end{table*}
\begin{table*}[ttt]
    \centering
    \caption{Overview of AUC in our methods and baselines with variances. The best and worst results per dataset are highlighted by blue and red, respectively. DNF stands for `Did Not Finish within 48 hours,' and OOM refers to `Out of Memory.' }
    \label{tab:overview_auc}
  %  \vspace{-2mm}
  \begin{minipage}[t]{1.0\linewidth}
  \centering
  \centering
  %\vspace{-2mm}
    \scalebox{0.95}{
    \begin{tabular}{crrrrrr|rr}\midrule
            &\multicolumn{8}{c}{{\bf Heterophilic}}\\
           &         {\bf Cornell} &              {\bf Wisconsin} &          {\bf Chameleon} &     {\bf Squirrel} &    {\bf Actor}&  \multicolumn{1}{r}{{\bf Penn94}} & \multicolumn{1}{|c}{Avg. }&\multicolumn{1}{c}{Rank}\\\midrule
           GraphNAS & \colorbox{redcolor}{86.1$_{\pm 0.2 }$}& \colorbox{redcolor}{89.8$_{\pm 0.1 }$}& 83.9$_{\pm 0.0 }$& 73.8$_{\pm 0.0 }$& \colorbox{redcolor}{65.1$_{\pm 0.0}$}& DNF &\colorbox{redcolor}{79.7}&\colorbox{redcolor}{4.2}\\
           GraphGym & 93.0$_{\pm 2.1 }$& 95.4$_{\pm 3.8}$& \colorbox{redcolor}{83.4$_{\pm 1.5}$} & \colorbox{redcolor}{71.4$_{\pm 1.0}$} & \colorbox{mycolor}{{\bf 72.4$_{\pm 0.6}$}}& \colorbox{mycolor}{{\bf 93.4$_{\pm 0.2}$}} &84.8&2.5\\
           DFG-NAS & \colorbox{mycolor}{{\bf 94.5$_{\pm 0.0}$}}& \colorbox{mycolor}{{\bf 95.8$_{\pm 0.0}$}}& 87.3$_{\pm 0.0}$ & 77.2$_{\pm 0.0 }$& 70.0$_{\pm 0.0 }$& \colorbox{redcolor}{84.8$_{\pm 0.0 }$}&84.9&\colorbox{mycolor}{{\bf 2.0}}\\
           Auto-HeG & 89.3$_{\pm 0.3 }$& \colorbox{mycolor}{{\bf 95.8$_{\pm 0.1}$}}& OOM & OOM & 70.0$_{\pm 0.0}$ & OOM &85.0&3.7\\\midrule
            \name &  91.0$_{\pm 0.1 }$& 93.6$_{\pm 0.2 }$& \colorbox{mycolor}{ \bf{87.6$_{\pm 0.0}$}}& \colorbox{mycolor}{{\bf 81.4$_{\pm 0.0}$}}& 70.1$_{\pm 0.0 }$& 91.9$_{\pm 0.0}$ & \colorbox{mycolor}{{\bf 85.9}}&2.2\\\midrule
    \end{tabular}
    }
\end{minipage}
\begin{minipage}[t]{1.0\linewidth}
  \centering
  \centering
  %\vspace{-2mm}
    \scalebox{0.95}{
    \begin{tabular}{crrrrrr|rr}\midrule
            &\multicolumn{8}{c}{{\bf Homophilic}}\\
           &   {\bf Cora} &         {\bf CiteSeer} &        {\bf Amz-P}  &   {\bf Amz-C} &         {\bf Co-CS} &        \multicolumn{1}{r}{{\bf PubMed}} & \multicolumn{1}{|c}{Avg.}&\multicolumn{1}{c}{Rank}\\\midrule
           GraphNAS & 97.2$_{\pm 0.0}$ & 92.1$_{\pm 0.0}$ & 99.5$_{\pm 0.0}$ & \colorbox{mycolor}{{\bf 99.2$_{\pm 0.0}$}} & \colorbox{redcolor}{99.7$_{\pm 0.0}$} & 97.0$_{\pm 0.0}$&97.5&2.7\\
           GraphGym & 96.7$_{\pm 0.5}$ & 92.6$_{\pm 0.6}$ & \colorbox{mycolor}{ \bf{99.6$_{\pm 0.2}$}} & \colorbox{mycolor}{{\bf 99.2$_{\pm 0.1}$}} & 99.8$_{\pm 0.0}$ & \colorbox{mycolor}{ \bf{97.5$_{\pm 0.3}$}}&97.6&\colorbox{mycolor}{{\bf 2.0}}\\
           DFG-NAS& \colorbox{mycolor}{{\bf 98.2$_{\pm 0.0}$}} & \colorbox{mycolor}{ \bf{93.6$_{\pm 0.0}$}} & 99.4$_{\pm 0.0}$ & 98.9$_{\pm 0.0}$ & \colorbox{mycolor}{ \bf{99.9$_{\pm 0.0}$}} & \colorbox{redcolor}{96.8$_{\pm 0.0}$}&\colorbox{mycolor}{{\bf 97.8}}&2.2\\
           Auto-HeG& \colorbox{redcolor}{92.9$_{\pm 0.0}$} & \colorbox{redcolor}{91.0$_{\pm 0.0}$} & OOM & OOM & OOM & \colorbox{redcolor}{96.8$_{\pm 0.0}$}&\colorbox{redcolor}{93.6}&\colorbox{redcolor}{4.8}\\\midrule
            \name & 98.0$_{\pm 0.0}$ & {93.4$_{\pm 0.0}$} & \colorbox{redcolor}{99.1$_{\pm 0.0}$} & \colorbox{redcolor}{98.8$_{\pm 0.0}$} & \colorbox{redcolor}{{99.7$_{\pm 0.0}$}} & 97.3$_{\pm 0.0}$&97.7&2.8\\\midrule
    \end{tabular}
    }
\end{minipage}
\end{table*}

\noindent
{\bf Baselines}.
We use four Graph NAS methods that codes are publicly open; GraphNAS~\cite{ijcai2020p195}, GraphGym~\cite{you2020design}, DFG-NAS~\cite{zhang2022deep}, and Auto-HeG~\cite{zheng2023auto}.
GraphNAS, DFG-NAS, and Auto-HeG use deep reinforcement learning, evolutionary algorithms, and differentiable search, respectively.
Since GraphGym does not have a search algorithm, we use uniform sampling following existing works~\cite{zhang2022deep}.

\noindent
{\bf Performance evaluation}.
We report the performance as average accuracy with their variances over five random runs.
We divide a set of nodes into train/validation/test in 0.6/0.2/0.2 following existing works~(e.g., \cite{zheng2023auto,chien2021adaptive}).
We also report run time to build GNN models and model sizes of the found models. 

\noindent
{\bf Hyper-parameters}.
We search for 1,000 GNN models following existing studies (e.g., \cite{wang2020neural,ijcai2020p195}).
For the hyper-parameters of \name, we set $\theta$ and $c$ as 10 and $\sqrt{2}$, respectively.
In GraphNAS, GraphGym, DFG-NAS, and Auto-HeG, we use default hyper-parameters provided at their GitHub repository. We do not tune hyper-parameters in Graph NAS methods to fairly compare run time, but DFG-NAS and Auto-HeG have specific hyper-parameters for datasets. We use data-specific hyper-parameters for DFG-NAS and Auto-HeG given by authors, so they might be over-tuned compared to others. 
We show details of hyper-parameter settings in our code.

\subsection{Performance comparison}
\label{sssec:effectiveness}
 
\noindent
{\bf Classification performance}.
Tables~\ref{tab:overview_acc} and \ref{tab:overview_auc} show the accuracy and AUC of each method, respectively. 
\name achieves high average accuracy and AUC in both heterophilic and homophilic graphs.
Interestingly, \name does not often achieve the highest and lowest performance; even if it has the lowest performance, the gaps are very small.
Existing Graph NAS methods, except for GraphNAS in heterophilic graphs and Auto-HeG in homophilic graphs, have averagely high performance.
The performance gap of GraphNAS between homophilic and heterophilic graphs indicates that the search space designed for homophilic graphs may not be suitable for heterophilic graphs.
Although GraphGym and DFG-NAS\footnote{Recall that we used data-specific hyper-parameters for DFG-NAS. Thus, DFG-NAS tends to have a good performance compared with other methods that have no data-specific hyper-parameters.} achieve the highest performance in some graphs, they have poor performance in some heterophilic graphs.
Auto-HeG has poor performance for homophilic graphs because it mainly focuses on heterophilic graphs.
Also, Auto-HeG is not scalable due to complex (i.e., large memory-consuming) GNN architectures, so it does not work in graphs with a large number of edges.
Consequently, \name achieves the highest average accuracy and AUC among twelve graphs, so we validated that \name{} is highly adaptable for both heterophilic and homophilic graphs.

\noindent
{\bf Model size}.
Table~\ref{tab:modelsize} shows the numbers of parameters on heterophilic graphs.
The numbers of parameters in the GNN architectures generated by \name are the smallest in all heterophilic graphs. 
In particular, in Chameleon and Squirrel, \name achieves the smallest models with the highest performance.
Since the search space of \name contains simple architectures, the found model has a small size.
These succinct architectures also contribute to interpretability.

% \begin{table*}[ttt]
%     \centering
%     \caption{Overview of model sizes }
%     \label{tab:modelsize}
%     %\vspace{-2mm}
%   \centering
%   %\vspace{-2mm}
%     \scalebox{1.0}{
%     \begin{tabular}{crrrrrr}\midrule
%             &\multicolumn{6}{c}{{\bf Heterophilic}}\\
%            &         {\bf Cornell} &              {\bf Wisconsin} &          {\bf Chameleon} &     {\bf Squirrel} &    {\bf Actor}&  \multicolumn{1}{r}{{\bf Penn94}}\\\midrule
%            GraphNAS & 1288.4K & 2214.7K & 1806.0K & 1584.9K & 1470.4K & OOM \\
%            GraphGym & 491.0K & 491.0K & 572.4K & 541.5K & 390.0K & 891.9K \\
%            DFG-NAS & 430.1K & 459.9K & 314.9K & 284.7K & 272.0K & 611.2K\\
%            Auto-HeG & 357.7K & 442.5K & OOM & OOM & 529.3K & OOM \\\midrule
%             \name & \colorbox{mycolor}{136.5K} & \colorbox{mycolor}{377.0K} & \colorbox{mycolor}{100.5K} & \colorbox{mycolor}{241.6K} & \colorbox{mycolor}{53.3K} & \colorbox{mycolor}{102.2K} \\\midrule
%     \end{tabular}
%     }
% \end{table*}

\begin{table*}[ttt]
    \centering
    \caption{Overview of model sizes }
    \label{tab:modelsize}
    %\vspace{-2mm}
  \centering
  %\vspace{-2mm}
    \scalebox{0.9}{
    \begin{tabular}{crrrrrr|rrrrrr}\midrule
            &\multicolumn{6}{c|}{{\bf Heterophilic}}&\multicolumn{6}{c}{{\bf Homophilic}}\\
           &         {\bf Cornell} &              {\bf Wisconsin} &          {\bf Chameleon} &     {\bf Squirrel} &    {\bf Actor}&  \multicolumn{1}{r|}{{\bf Penn94}} &        {\bf Cora} &         {\bf CiteSeer} &        {\bf Amz-P}  &   {\bf Amz-C} &         {\bf Co-CS} &        \multicolumn{1}{r}{{\bf PubMed}}\\\midrule
           GraphNAS & 1288.4K & 2214.7K & 1806.0K & 1584.9K & 1470.4K & OOM & 1590.7K & 1213.1K & 224.6K & 239.7K & 1346.3K & 320.1K \\
           GraphGym & 491.0K & 491.0K & 572.4K & 541.5K & 390.0K & 891.9K & 456.1K & 753.2K & 366.3K & 369.7K & 1161.6K & 332.9K \\
           DFG-NAS & 430.1K & 459.9K & 314.9K & 284.7K & 272.0K & 611.2K & \colorbox{mycolor}{191.1K} & 1699.8K & \colorbox{mycolor}{96.5K} & \colorbox{mycolor}{99.6K} & \colorbox{mycolor}{873.1K} & 94.2K \\
           Auto-HeG & 357.7K & 442.5K & OOM & OOM & 529.3K & OOM & 667.3K & 930.2K & OOM & OOM & OOM & 69.7K \\\midrule
            \name & \colorbox{mycolor}{136.5K} & \colorbox{mycolor}{377.0K} & \colorbox{mycolor}{100.5K} & \colorbox{mycolor}{241.6K} & \colorbox{mycolor}{53.3K} & \colorbox{mycolor}{102.2K} & 317.1K & \colorbox{mycolor}{545.6K} & 143.5K & 111.0K & 1812.8K & \colorbox{mycolor}{69.6K} \\\midrule
    \end{tabular}
    }
    
\end{table*}

\begin{figure}[!t]
    \begin{minipage}[t]{1.0\linewidth}
    \includegraphics[width=1.0\linewidth]{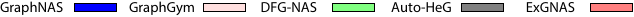}
    \end{minipage}
    \begin{minipage}[t]{1.0\linewidth}
    \centering
    \includegraphics[width=1.0\linewidth]{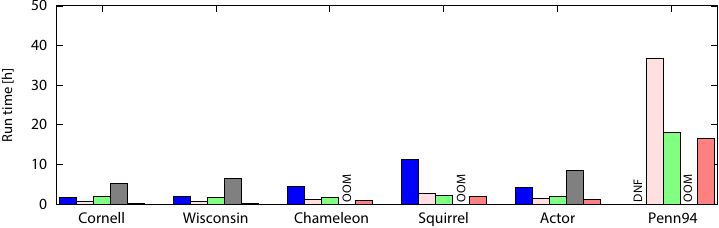}
    \subcaption{Heterophilic graphs}
     \end{minipage}
    \begin{minipage}[t]{1.0\linewidth}
    \centering
    \includegraphics[width=1.0\linewidth]{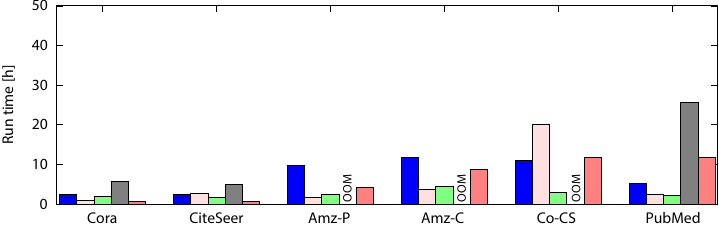}
        \subcaption{Homophilic graphs}
     \end{minipage}
\caption{Run time [h] including search and model training time.  We terminated the run if it did not finish within 48 hours.  OOM indicates out-of-memory in Auto-HeG. }\label{fig:runtime}
\end{figure}

\noindent
{\bf Search efficiency}.
Figure~\ref{fig:runtime} shows the run time of Graph NAS methods in homophilic and heterophilic graphs.
The run time includes the architecture search and model training time of GNNs.
From these results, \name, GraphGym, and DFG-NAS are efficient because their search algorithms do not use neural models. 
\name is much more efficient than the other methods in heterophilic graphs, while GraphGym and DFG-NAS are often faster than \name in homophilic graphs.
This indicates that the search space has impacts on model training time as well as accuracy.
GraphNAS and Auto-HeG take a long time because their search algorithms use deep learning.
We can validate that our simple search space and algorithm can efficiently find accurate GNN models.

\noindent
{\bf Summary}.
While \name shows only modest improvement in each individual metric over the baselines, the fact that accuracy, efficiency, and model size all improve simultaneously is remarkable.
For example, compared with DFG-NAS, \name achieves an average improvement of 0.5 in ACC and 0.45 in AUC, along with a 31\% reduction in run time and a 224\% reduction in model size.

% \begin{table}[ttt]
%     \centering
%     \caption{Search algorithm difference.}
%    \label{tab:acc_searchalgorithm}
%     \vspace{-3mm}
%     \scalebox{0.8}{
%     \begin{tabular}{crr}\midrule
%         &\multicolumn{1}{c}{{\bf Heterophilic}}&\multicolumn{1}{c}{{\bf Homophilic}}\\\hline
%         HomGraphGym&84.6&97.6\\
%         HeteroUniform&85.5 &97.3 \\
%         Max&85.4 &97.3  \\
%         \name &\colorbox{mycolor}{85.9}&\colorbox{mycolor}{97.6}\\\midrule        
%     \end{tabular}
%     }
% \end{table}

\begin{table}[ttt]
    \centering
    \caption{Comparison of search spaces and algorithms.}
   \label{tab:acc_searchalgorithm}
    %\vspace{-3mm}
    \scalebox{1.0}{
    \begin{tabular}{c|rrrr}\midrule
        %&\multicolumn{3}{c|}{{\bf Heterophilic}}&\multicolumn{3}{c}{{\bf Homophilic}}\\
           &   {\bf GraphGym} &    {\bf Uniform}&         {\bf Max} &   {\bf \name} \\\midrule
        % GCN&195.7 &259.9 &194.1 &52.0 &54.1 &672.6 \\
        % SGC&8.5 &8.5 &11.6 &10.5 &4.7 &9.5 \\
        % GAT&886.3 &632.0 &1440.8 &381.7 &950.5 &337.5 \\
        % GCNJ&358.3 &248.7 &1020.2 &1810.9 &226.3 &123.9 \\
        % GATJ&385.5 &476.2 &797.7 &710.6 &39.8 &315.4 \\
        % APPNP&617.2 &845.1 &786.1 &1190.5 &707.3 &2372.7 \\
        % MIXHOP&264.4 &503.2 &367.5 &619.2 &401.4 &851.1 \\
        Homophilic    &97.6   & 97.3 &  97.3  & \colorbox{mycolor}{97.7}  \\
        Heterophilic  &84.8   & 85.5 &  85.4  & \colorbox{mycolor}{85.9}  \\\midrule
        % H2GCN&872.0 &273.7 &277.3 &161.7 &2286.0 &121.0 \\
        % LINX&1921.6 &5053.5 &23370.8 &844.0 &6689.6 &1175.5 \\
        % GloGNN++&950.5 &1109.5 &5596.2 &1876.5 &3236.4 &2604.9 \\\midrule
        % % GraphNAS&221.2 &230.9 &619 &136.6 &31.1 &---\\
        % % \name (max)&273.3 &366.6 &255.9 &138.6 &112.7 &553.2 \\
        % \name &\colorbox{mycolor}{138.6}&\colorbox{mycolor}{176.4}&367.8 &\colorbox{mycolor}{83.0}&\colorbox{mycolor}{804.5}&\colorbox{mycolor}{76.1}\\\midrule        
    \end{tabular}
    }
\end{table}

\begin{figure}[t]
\begin{minipage}[t]{0.48\linewidth}
    \centering
	\includegraphics[width=0.9\linewidth]{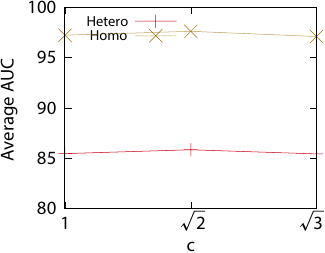}
        \subcaption{$c$}
\end{minipage}
\hfill
\begin{minipage}[t]{0.48\linewidth}
	\includegraphics[width=0.9\linewidth]{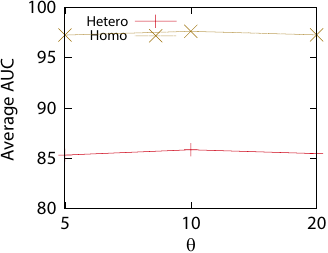}
         \subcaption{$\theta$}
\end{minipage}
\caption{Impact on hyper-parameters $c$ and $\theta$}\label{fig:hyperparameter}
\end{figure}

\subsection{Analysis of \name}

{\bf Comparison of search space/algorithm}.
Table~\ref{tab:acc_searchalgorithm} shows the AUC on GraphGym, uniform sampling with our search space (Uniform), and preferential search for the maximum accuracy with our search space (Max).
We can compare the performance difference between the search spaces of \name{} and GraphGym and between the Monte-Carlo tree search and other search algorithms.
First, Uniform has a higher AUC than GraphGym in heterophilic graphs, while GraphGym has a higher AUC than Uniform in homophilic graphs.  
This indicates that our search space is more suitable for heterophilic graphs than that of GraphGym because they use the same search algorithm.
\name outperforms Uniform and Max, so we can confirm that Monte-Carlo tree search preferentially searches for sub-optimal architectures.

\begin{figure*}[ttt]
  \begin{minipage}{1.0\textwidth}
  \center
	\includegraphics[width=0.7\linewidth]{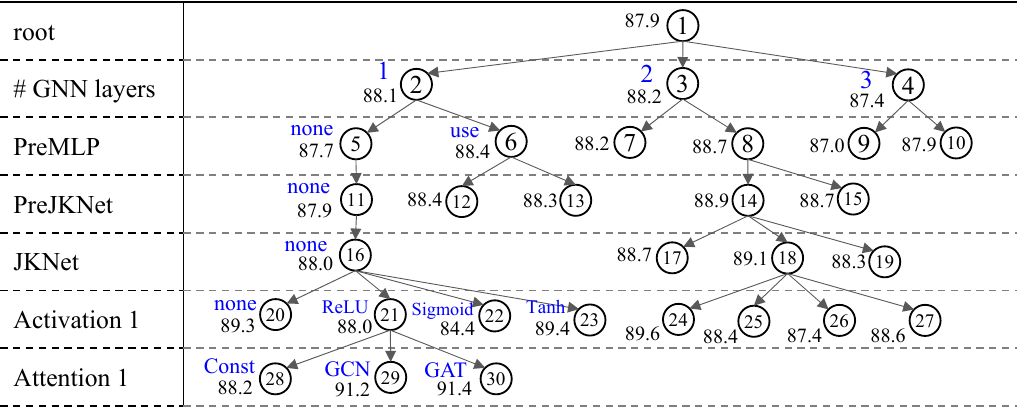}
  \subcaption{CiteSeer}
  \end{minipage}
  \hfill
  \begin{minipage}{1.0\textwidth}
    \center
	\includegraphics[width=0.7\linewidth]{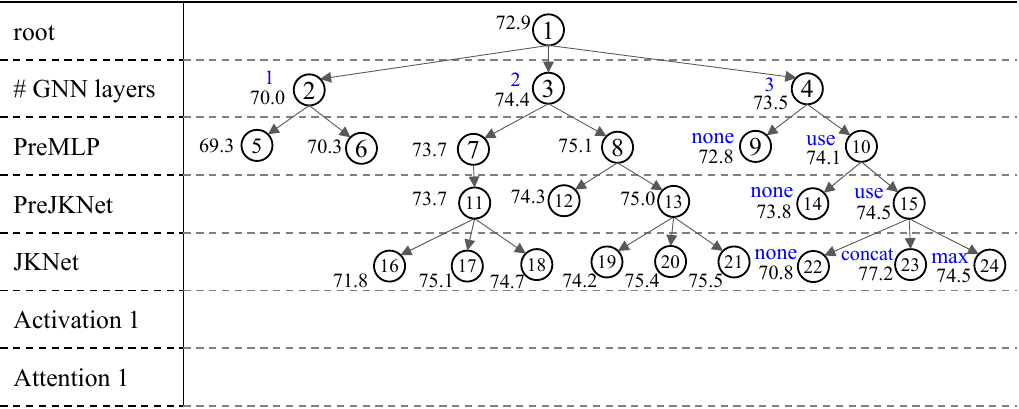}
  \subcaption{Squirrel}
  \end{minipage}
\caption{Examples of Monte-Carlo trees. Blue-colored letters indicate the architecture parameters. We note that we omit unnecessary parts for discussions.}
\label{fig:montecarlotreeoutputs}
\end{figure*}

\noindent
{\bf Hyper-parameter sensitivity}.
\name has two hyper-parameters, $c$ and $\theta$. Figure~\ref{fig:hyperparameter} shows their impact on the AUC.
This result shows that $c$ and $\theta$ are not sensitive to the classification performance.
Therefore, we can use default parameters without tremendous hyperparameter turnings, which leads to reducing computational costs and improving ease of use.

\subsection{Explainability and architecture analysis}
\label{sssec:architecturenalysis}

We show how \name{} is helpful to analyze GNN architectures. We focus on the difference between GNN architectures for homophilic and heterophilic graphs.  
We note that these analyses cannot be conducted by other Graph NAS because they cannot find suitable GNN architecture for both homophilic and heterophilic graphs.

\begin{figure*}[t]
\begin{minipage}[t]{0.24\linewidth}
    \centering
	\includegraphics[width=1.0\linewidth]{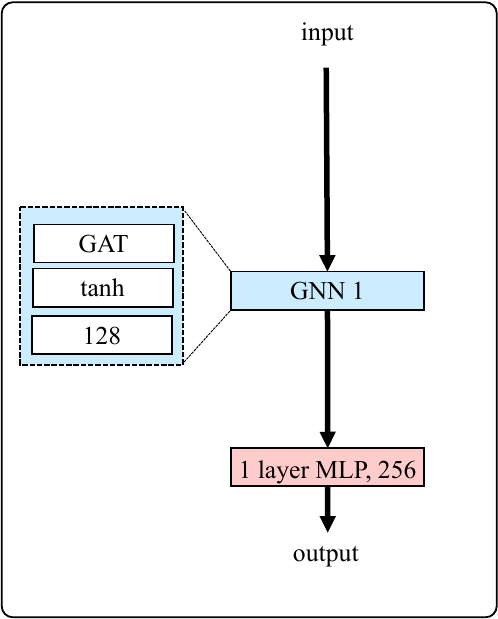}
 \subcaption{CiteSeer}
\end{minipage}
\begin{minipage}[t]{0.24\linewidth}
	\includegraphics[width=1.0\linewidth]{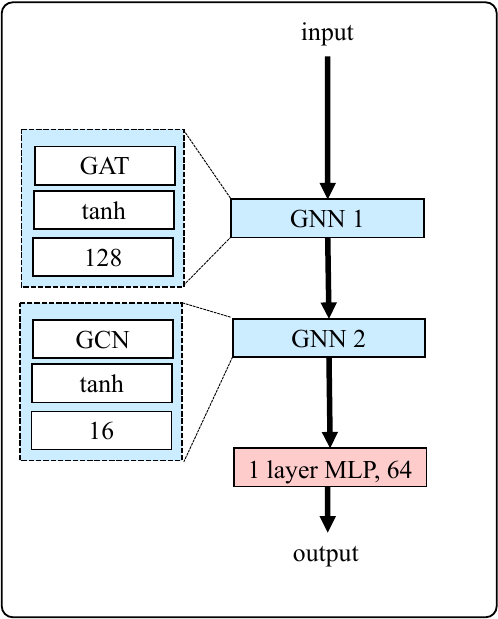}
 \subcaption{Amz-c}
\end{minipage}
\begin{minipage}[t]{0.24\linewidth}
	\includegraphics[width=1.0\linewidth]{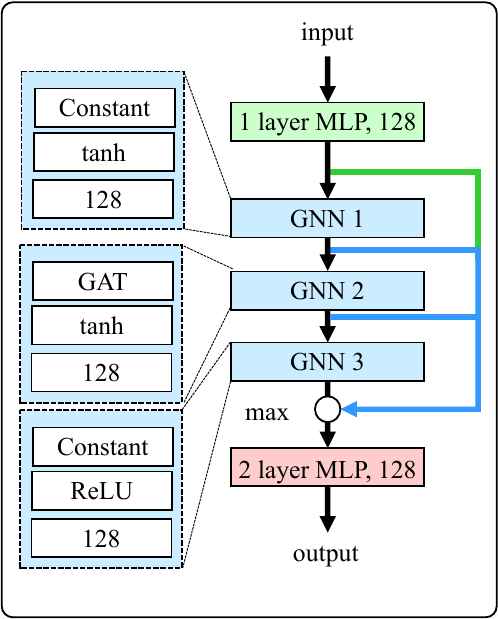}
 \subcaption{Cornell}
\end{minipage}
\begin{minipage}[t]{0.24\linewidth}
\includegraphics[width=1.0\linewidth]{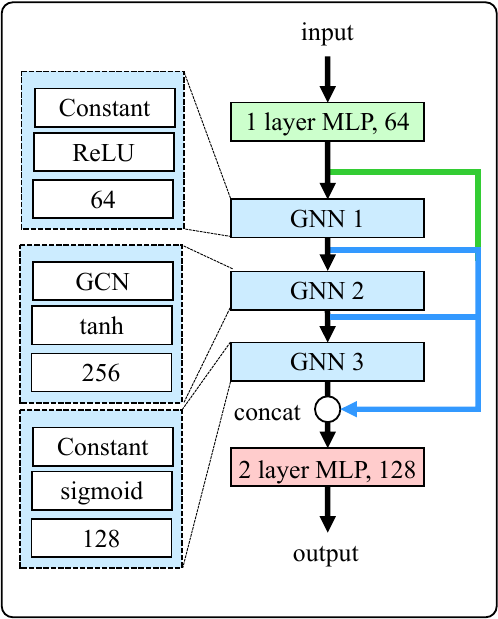}
\subcaption{Squirrel}
\end{minipage}
\caption{Selected GNN architectures; CiteSeer and Amz-c are homophilic graphs and Cornell and Squirrel are heterophilic graphs.}\label{fig:bestarchitecture}
\end{figure*}

\noindent
{\bf Examples of Monte-Carlo trees}. 
Figure~\ref{fig:montecarlotreeoutputs} illustrates the Monte-Carlo trees for CiteSeer and Squirrel corresponding to GNN architectures in Figure~\ref{fig:bestarchitecture}.
Each value next to MCT nodes is the average AUC of GNN architectures.
We can know important components and effective/ineffective combinations of architecture parameters from these trees.
In CiteSeer, the average AUC does not have a large gap between the number of layers one and two (i.e., MCT nodes 2 and 3). However, if we select a single GNN layer, ReLU activation function, and GCN/GAT (i.e., MCT node 29, 30), the AUC becomes high. 
If we select a constant for attention (i.e., MCT node 28), the AUC does not become high.
In addition, if we select a single GNN layer and Sigmoid activation function (i.e., MCT node 22), the AUC becomes quite low. This example shows that combinations of architecture parameters are important rather than architecture parameters themselves.

In Squirrel, if we select concat JKNet (i.e., MCT nodes 17, 20, and 23), the average AUC is high, which indicates concat JKNet is averagely effective for Squirrel.
In addition, if the GNN architecture includes three GNN layers and concat JKNet (i.e., MCT node 23), the average AUC becomes quite high. 
This indicates that it is effective to separately aggregate embeddings from high-order neighborhoods.  
We can analyze the importance of architecture parameters and their combinations from Monte-Carlo trees.

\noindent
{\bf Selected architectures}.
We here show the difference between architectures in homophilic and heterophilic graphs. 
Figure~\ref{fig:bestarchitecture} illustrates examples of GNN architectures that \name{} found in CiteSeer, Amz-c, Cornell, and Squirrel.
These examples show that the best architectures significantly differ across graphs.
The GNN architectures for heterophilic graphs (i.e., Cornell and Squirrel) are more complex than those for homophilic graphs (i.e., CiteSeer and Amz-c).
The best architecture for CiteSeer is quite simple, so it indicates that labels can be predicted from attributes of neighborhoods.
In Cornell and Squirrel, attributes of own and high-order neighborhoods are important.

Table~\ref{tab:ratio_functions} shows the ratios of selected architecture parameters in our experiments.
This result reveals four interesting insights.
First, GNN architectures for heterophilic graphs should stack more multiple GNN layers than ones for homophilic graphs. The default number of layers in most GNNs is two but it may not be optimal for heterophilic graphs.
Second, Hyperbolic tangent is often selected as an activation function in the best architectures for both homophilic and heterophilic graphs, though ReLU is commonly used for GNN architectures. 
Our results suggest Hyperbolic tangent is better than ReLU. Also, for homophilic graphs, no activation functions often work well.
Finally, the gap is small between ratios of selected attention functions in homophilic and heterophilic graphs, so it may be hard to decide the best attention functions among concat, GCN, and GAT from graph types.
These insights help to design GNN architectures.
 
\begin{table*}[ttt]
    \centering
    \caption{The ratio of the selected architectural parameters}
    \label{tab:ratio_functions}
    %\vspace{-3mm}
    \scalebox{1.0}{
    \begin{tabular}{c|rrr|rrrr|rrr}\hline
        &\multicolumn{3}{|c|}{{\bf \# of GNN layers}} & \multicolumn{4}{|c|}{{\bf Activation}} &\multicolumn{3}{|c}{{\bf JKNet}} \\
        & 1 & 2 & 3 & none & ReLU & Sigmoid & Tanh& none & concat & max 
        \\\hline
    Hom & 0.14 & 0.63 & 0.23 & 0.32 & 0.10 & 0.04 & 0.55  & 0.29 & 0.54 & 0.17
    \\
    Hetero & 0 & 0.47 & 0.53 & 0.11 & 0.21 & 0.17 & 0.51 & 0.03 & 0.67 & 0.30
    \\\hline
    \end{tabular}
    }
\end{table*}

\subsection{ User study on effectiveness of Monte-Carlo trees on explainability}
We conduct a questionnaire-based user study for Monte-Carlo trees to validate the effectiveness of explainability of \name.
The first and second questions for GNN developers are (1) Is the Monte-Carlo tree useful to develop new GNN architectures and (2) Is the Monte-Carlo tree useful to explore the best GNN when you apply GNNs to your service, respectively.
Each participant answers five-level Likert-scale questions (i.e., very good to very poor) for each question and comments, with viewing Figure~\ref{fig:montecarlotreeoutputs} and its explanations.

We received $16$ answers from participants who do research/develop/use GNNs in various fields such as computer/data science, material science, and medicine.
For the first question, we received 7 very good and 9 good; all answers are positive.
For example, users answered that ``It facilitates the analysis of which factors contribute to optimal predictions, leading to improved model interpretability.'', ''This method facilitates an intuitive understanding of the impact of effective network architectures on accuracy.'', and ``The optimal structure can be visually identified, it becomes easier for those handling the output to determine which GNN architecture to use, and it also allows others to easily understand the rationale behind choosing that particular structure.''

For the second question, we received 11 very good, 2 good, 2 fair, and 1 poor; more than 80\% users are positive.
Example of positive comments is ``Once we understand what the key components are, we can determine which search space we should focus on for further performance improvement.''
The negative comment is  ``It is likely that in many cases, models are selected through benchmarks for well-established GNNs. In such situations, it is difficult to consider the internal workings of the GNN in detail.''

We showed empirical evidence that the Monte-Carlo tree is useful for many developers in both scenarios, which were not validated in prior studies.

\subsection{Re-designing GCN}
Our Monte-Carlo tree helps to analyze the effective functions.
In our results, Hyperbolic tangent is often selected as an activation function in the best architectures for both homophilic and heterophilic graphs, though ReLU is commonly used for GNN architectures. 
Thus, we here evaluate the performance of GCN with Tanh and ReLU functions.

\begin{table*}[h]
    \centering
    \caption{Impact on ReLU and Tanh in GCN}
   \label{tab:impactgcn}
    %\vspace{-3mm}
    \scalebox{1.0}{
    \begin{tabular}{c|rrrrrr}\hline
          &   {\bf Cornell} &         {\bf Wisconsin} &        {\bf Chameleon}  &   {\bf Squirrel} &    {\bf Actor}&         {\bf Penn94}\\\hline
        % GCN&195.7 &259.9 &194.1 &52.0 &54.1 &672.6 \\
        % SGC&8.5 &8.5 &11.6 &10.5 &4.7 &9.5 \\
        % GAT&886.3 &632.0 &1440.8 &381.7 &950.5 &337.5 \\
        % GCNJ&358.3 &248.7 &1020.2 &1810.9 &226.3 &123.9 \\
        % GATJ&385.5 &476.2 &797.7 &710.6 &39.8 &315.4 \\
        % APPNP&617.2 &845.1 &786.1 &1190.5 &707.3 &2372.7 \\
        % MIXHOP&264.4 &503.2 &367.5 &619.2 &401.4 &851.1 \\
        GCN with ReLU& 70.2 & 67.6 & 85.7 & 71.9 & 56.4 & 89.3 \\
        GCN with Tanh& 71.1 & 73.5 & 80.3 & 69.8 & 58.1 & 87.9 \\\hline
        % GraphNAS&221.2 &230.9 &619 &136.6 &31.1 &---\\
        % \name (max)&273.3 &366.6 &255.9 &138.6 &112.7 &553.2 \\
    \end{tabular}
    }
    %\vspace{-5mm}
\end{table*}

Table~\ref{tab:impactgcn} shows the AUC of GCNs with ReLU and Tanh in heterophilic graphs.
In Cornell, Wisconsin, and Actor, the AUC increased; in particular, the AUC increased by 6.0 in Wisconsin.
This experiment suggests that we need to carefully select activation functions depending on graphs.
We validate that \name helps to analyze the important components and re-design existing GNN architectures.

\section{Conclusion}
\label{sec:conclusion}

We introduced the explainable graph neural architecture search problem, which aims to output the best GNN model and the importance of its components. 
We proposed an efficient and explainable Graph NAS method via Monte-Carlo tree search, called \name, that can handle both homophilic and heterophilic graphs.
Our experimental study showed that \name{} (i) effectively and efficiently finds models that achieve high accuracy compared with Graph NAS method and (ii) helps to analyze the GNN architecture. 

In the future, we plan to (1) extend our method to automatically generate/order architecture parameters according to graphs, (2) explore more sophisticated search space and search algorithms, and (3) evaluate our method in other graph types and tasks such as directed graphs, link prediction, and fairness~\cite{sasaki2025benchmarking,laclau2022survey,pechenizkiy2025benchmarking}.

% \section*{Acknowledgement}
% This work was supported by Japan Science and Technology Agency (JST) as part of Adopting Sustainable Partnerships for Innovative Research Ecosystem (ASPIRE), Grant Number JPMJAP2328, and JST Presto Grant Number JPMJPR21C5.

\bibliography{graphassociationrule}
\bibliographystyle{plain}

% \appendix
% \input{99-appendix}
\end{document}